\documentclass[runningheads]{llncs}

\usepackage{eccv}

\usepackage{eccvabbrv}

\usepackage{graphicx}
\usepackage{booktabs}

\usepackage[accsupp]{axessibility}  

\usepackage{hyperref}

\usepackage{orcidlink}

\begin{document}

\title{RAE-AR: Taming Autoregressive Models with Representation Autoencoders} 

\titlerunning{Abbreviated paper title}


\author{Hu Yu\inst{1,2}$^{*}$ \and
Hang Xu\inst{1,2}$^{*}$ \and
Jie Huang\inst{2} \and
Zeyue Xue\inst{2,3} \and
Haoyang Huang\inst{2} \and \\
Nan Duan\inst{2} \and
Feng Zhao\inst{1} }
\authorrunning{Hu Yu et al.}

\institute{University of Science and Technology of China \and
JD Explore Academy \and The University of Hong Kong \\
\url{https://yuhu98.github.io//projects/RAE-AR.html}
}

\maketitle
\begingroup
\renewcommand\thefootnote{}\footnotetext{* Equal contribution.}
\endgroup

\begin{abstract}
  The latent space of generative modeling is long dominated by the VAE encoder. The latents from the  pretrained representation encoders (e.g., DINO, SigLIP, MAE) are previously considered inappropriate for generative modeling. Recently, RAE method lights the hope and reveals that the representation autoencoder can also achieve competitive performance as the VAE encoder. However, the integration of representation autoencoder into continuous autoregressive (AR) models, remains largely unexplored.
  In this work, we investigate the challenges of employing high-dimensional representation autoencoders within the AR paradigm, denoted as \textit{RAE-AR}.
  We focus on the unique properties of AR models and identify two primary hurdles: complex token-wise distribution modeling and the high-dimensionality amplified training-inference gap (exposure bias).
  To address these, we introduce token simplification via distribution normalization to ease modeling difficulty and improve convergence. Furthermore, we enhance prediction robustness by incorporating Gaussian noise injection during training to mitigate exposure bias. Our empirical results demonstrate that these modifications substantially bridge the performance gap, enabling representation autoencoder to achieve results comparable to traditional VAEs on AR models. This work paves the way for a more unified architecture across visual understanding and generative modeling.

  \keywords{Representation Autoencoder \and Autoregressive Models}
\end{abstract}

\section{Introduction}
\label{sec:intro}

In recent years, Variational Autoencoders (VAEs) trained via reconstruction have dominated the generative landscape, serving as the foundation for both diffusion models \cite{rombach2022high} and autoregressive frameworks \cite{li2024autoregressive}. While generative modeling has evolved significantly through advancements in diffusion backbones and autoregressive strategies, the underlying autoencoders that define the latent space have remained largely stagnant. 
Concurrently, the field of visual representation learning has undergone a rapid transformation \cite{radford2021learning, he2022masked, oquab2023dinov2, tschannen2025siglip, assran2023self}. 
Although some recent studies have attempted to enhance latent quality indirectly, for instance, through REPA-style \cite{yu2024representation} alignment with external encoders, the direct integration of representation autoencoder into generative modeling has long been considered a significant challenge.

\begin{figure}[t]
    \begin{center}
	\includegraphics[width=0.98\linewidth]{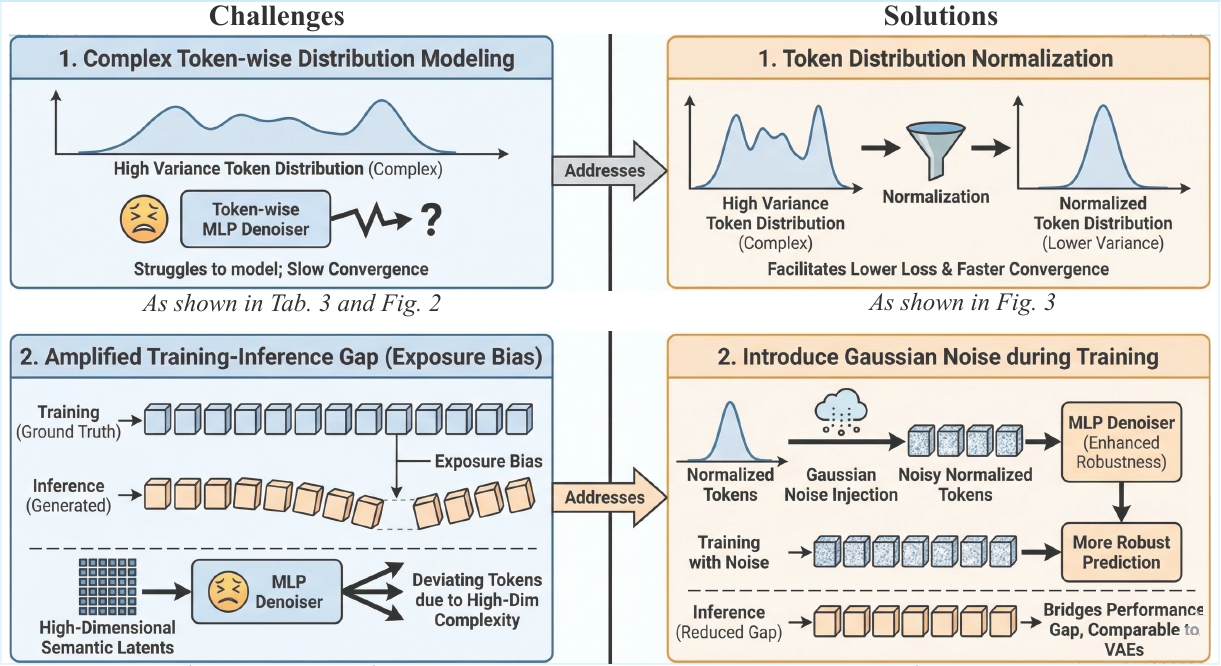}
    \end{center}
    \setlength{\abovecaptionskip}{-0.2cm}
    \setlength{\belowcaptionskip}{-0.2cm}
    \caption{The challenges of integrating representation autoencoders into AR models: token variance and exposure bias. We correspondingly take token normalization and noise perturbation strategies.}
    \label{fig:main}
    \vspace{-3mm}
\end{figure}

Recently, several pioneering works \cite{zheng2025diffusion, shi2025latent, tong2026scaling, shi2025svg} have achieved significant breakthroughs in the use of semantic latents. These studies demonstrate that high-dimensional, semantic-based visual encoders can serve as powerful latent spaces for diffusion models, delivering performance competitive with traditional reconstruction-based VAEs. For instance, RAE \cite{zheng2025diffusion} introduces specialized modifications by increasing model width and optimizing the noise schedule, and SVG \cite{shi2025latent} explicitly enhances the latent representation through a residual branch architecture.
Collectively, these advancements highlight the immense potential for unifying visual understanding and generation. However, despite these gains in the diffusion domain, the compatibility and efficacy of semantic representations within autoregressive models remain largely unexplored.

The emergence of continuous autoregressive models \cite{li2024autoregressive, fan2024fluid, yu2025frequency} has enabled the autoregressive paradigm to utilize the same latent representations as diffusion models. Consequently, previous works have largely adopted the standard reconstruction-based VAEs popularized by diffusion frameworks. To investigate this further, we first evaluate the performance of various reconstruction-based and semantic-based latents within the autoregressive context, including VAE \cite{kingma2013auto}, VA-VAE \cite{yao2025reconstruction}, DINOv2 \cite{oquab2023dinov2}, SigLIP2 \cite{tschannen2025siglip}, and MAE \cite{he2022masked}. Our initial findings indicate that semantic-based latents yield significantly inferior results compared to their reconstruction-based counterparts. This observation mirrors a known challenge in diffusion modeling, where the direct integration of semantic encoders often results in degraded generative quality. These findings raise a critical research question: is it possible to effectively adapt semantic encoders to function within the autoregressive generative paradigm?

In this paper, we investigate the challenges of integrating high-dimensional representation autoencoders within the autoregressive (AR) paradigm, a framework we denote as RAE-AR. We deeply investigate the unique properties of AR models, identifying two primary challenges that may influence the performance: complex token-wise distribution modeling and the amplified training-inference gap (exposure bias) resulting from high-dimensionality, as shown in the left part of Fig. \ref{fig:main}. Regarding the former, whereas diffusion models learn image-wide distributions via bidirectional attention, continuous AR models learn token-wise distributions through an MLP-based denoiser. By analyzing the token distributions of existing latent representations, we reveal that a lower token-wise distribution variance typically correlates with superior performance. Consequently, we propose token distribution normalization for representation autoencoders, which facilitates lower training loss and accelerated convergence (though, as discussed later, this does not surely guarantee improved final synthesis quality due to exposure bias).
Furthermore, the inherent training-inference gap, where generated tokens deviate from the ground-truth distribution, represents a distinct hurdle for AR models. This exposure bias is further exacerbated by the high dimensionality and large information capacity of semantic latents, since denoising MLP struggles to accurately model such complex, high-dimensional distributions. To address this, we enhance prediction robustness by introducing Gaussian noise to the normalized tokens during training. Empirical results demonstrate that these modifications substantially bridge the performance gap, enabling representation autoencoders to achieve results comparable to VAEs.

\section{Related Works}

\subsection{Representation for Reconstruction and Generation.}
Visual representation learning is evolving at breakneck speed. Self-supervised and multimodal encoders such as DINO \cite{oquab2023dinov2}, MAE \cite{he2022masked}, JEPA \cite{assran2023self} and CLIP \cite{radford2021learning} learn semantically structured visual features that generalize across tasks and scales and provide a natural basis for visual understanding. However, latent diffusion models remain largely isolated from this progress, continuing to operate within reconstruction-trained VAE spaces rather than semantically meaningful representational ones.
Later, some efforts explore enhancing VAEs with semantic representations. For instance, VA-VAE \cite{yao2025reconstruction} aligns VAE latents with a pretrained representation encoder. MAETok \cite{chen2025masked} and DC-AE 1.5 \cite{chen2025dc} incorporate MAE- or DAE \cite{vincent2008extracting}-inspired objectives into VAE training.

Contemporary research also explores the direct application of semantic representations to improve the generative modeling process itself. Approaches like REPA \cite{yu2024representation} accelerate the convergence of Diffusion Transformers (DiT) by aligning intermediate block features with those of external representation encoders.
Other methodologies \cite{yue2025uniflow, chen2025aligning} finetune the semantic encoders to capture finer details, but reduce the dimension to lower the difficulty. 
SVG \cite{shi2025latent, shi2025svg} adds a residual branch based on DINOv3 \cite{simeoni2025dinov3} for better details. 
RAE \cite{zheng2025diffusion, tong2026scaling, tong2026beyond} keeps the semantic encoder frozen but introduces complicated architectural and training modifications. However, despite these significant strides in the diffusion domain, the integration of semantic latents on autoregressive models are rarely explored.

\subsection{Autoregressive Image Generation}
Apart from diffusion models, autoregressive models also achieve great advancement. According to regression directions, autoregressive image generation can be grouped into several types: \textit{next-token}, \textit{next-set}, \textit{next-scale}, and \textit{next-frequency} prediction. 
The \textit{next-token} prediction \cite{sun2024autoregressive}, inspired by Large Language Models (LLMs) \cite{brown2020language}, employs a raster-scan method to flatten the interdependent 2-D latent tokens. 
The \textit{next-set prediction} (also denoted as masked autoregression) \cite{chang2022maskgit, li2024autoregressive}, derived from BERT \cite{devlin2018bert}, predicts the masked tokens given the unmasked ones. 
For the \textit{next-scale} prediction, VAR \cite{tian2024visual} combines RQ-VAE \cite{lee2022autoregressive} with multi-scale, aggregating all scales to produce the final prediction. 
For the \textit{next-frequency} prediction, FAR \cite{yu2025frequency} leverages the spectral dependency of image data, and proposes to generate from low frequency to high frequency. 

Apart from regression directions, another focus of autoregressive model is the latent representation. The mainstream way is discrete tokens with vector quantization \cite{esser2021taming, sun2024autoregressive}. Recently, some methods \cite{li2024autoregressive, fan2024fluid, yu2025frequency, yu2025videomar} reveal the possibility of employing continuous latents for overcoming the significant information quantization loss. For example, MAR \cite{li2024autoregressive} employs VAE as latent representation for image generation. VideoMAR \cite{{yu2025videomar}} adopts Cosmos-Tokenizer \cite{agarwal2025cosmos}, a video VAE for video generation. However, the exploration of semantic-based latents are rarely explored in the continuous autoregressive model paradigm.

\section{Preliminary}
\label{sec:Preliminaries}

\subsection{Diffusion Loss for Continuous Tokens}
For the tokenizer in autoregressive models, the key is to model the per-token probability distribution, which can be measured by a loss function for training and a token sampler for inference. Following MAR \cite{li2024autoregressive}, we adopt diffusion models to solve these two bottlenecks for integrating continuous tokenizer into autoregressive models.

\noindent {\bf Loss function.} Given a continuous token $z$ produced by a autoregressive transformer model and its corresponding ground-truth token $x$, MAR employs diffusion model as loss function, with $z$ being the condition. 
\begin{equation}
    \mathcal{L}(z, x)=\mathbb{E}_{\varepsilon, t}\left[\left\|\varepsilon-\varepsilon_{\theta}\left(x_{t} \mid t, z\right)\right\|^{2}\right] .
\end{equation}
Here, $\varepsilon \sim \mathcal{N}(\mathbf{0}, \boldsymbol{I})$, and $x_{t}=\alpha_{t} x_{0}+\sigma_{t} \varepsilon$, with $\alpha_{t}$ being the noise schedule \cite{ho2020denoising}. The noise estimator $\varepsilon_{\theta}$, parameterized by $\theta$, is a small MLP network.

\noindent {\bf Token sampler.} The sampling procedure totally follows the inference process of diffusion model. Starting from $x_T \sim \mathcal{N}(\mathbf{0}, \boldsymbol{I})$, the reverse diffusion model iteratively remove the noise and produces $x_0 \sim p(x|z) $, under the condition $z$.

\subsection{Representation Autoencoder}
For the reconstruction task, we follow the settings of RAE \cite{zheng2025diffusion}. We freeze the representation encoders to serve as the diffusion latent space, and pair it with a properly trained ViT-based decoder. 

The training recipe for the RAE decoder is described concisely as follows.
The RAE decoder $D$ is trained to reconstruct an input image $x \in \mathbb{R}^{3\times H\times W}$ from tokens $z$ generated by a frozen representation encoder $E$. For $256\times256$ images, the encoder produces $N = 256$ tokens with hidden dimension $d$, maintaining consistency with standard DiT-based models. By setting the decoder’s patch size equal to the encoder’s patch size, the model ensures the reconstructed output $\hat{x}$ matches the original input resolution. The optimization follows standard VAE practices, employing a weighted combination of L1, LPIPS \cite{zhang2018unreasonable}, and adversarial (GAN) \cite{goodfellow2014generative} losses:

\begin{equation}
    \mathcal{L}_{rec}(x) = \text{L1}(\hat{x}, x) + \omega_L \text{LPIPS}(\hat{x}, x) + \omega_G \lambda \text{GAN}(\hat{x}, x).
\end{equation}

To evaluate the framework across diverse pretraining paradigms, we employ three representative backbones as \textbf{experimental settings}: DINOv2-B (self-distillation), SigLIP-B (language supervised), and MAE-B (masked autoencoding). Unless otherwise noted, an ViT-XL decoder is used for all experiments.

\section{Analysis: Rec\&Gen Performance of RAE on AR Model}

\begin{table}[b]
    \setlength{\tabcolsep}{10pt}
    \centering
    \caption{The reconstruction performance of the five generative latents. The representation autoencoders have comparable reconstruction performance as the VAEs.}
    \label{tab:Reconstruction}
    \begin{tabular}{c|ccccc}
    \toprule
    Reconstruction  &  VAE  &  VA-VAE  &  DINOv2  &  SigLIP2  &  MAE  \\
    \midrule[0.1em]
    Metric & f16d16 & f16d32  &  f16d768  &  f16d768  &  f16d768  \\
    \midrule
    rFID$\downarrow$ & 0.533  & 0.279  & 0.619  & 0.703  & 0.643  \\
    PSNR$\uparrow$   & 26.184 & 27.705 & 19.202 & 19.655 & 29.441 \\
    LPIPS$\uparrow$  & 0.135  & 0.097  & 0.223  & 0.218  & 0.085  \\
    SSIM$\uparrow$   & 0.716  & 0.779  & 0.639  & 0.664  & 0.914  \\
    \bottomrule
    \end{tabular}
\end{table}

Existing latent representations can be broadly categorized into two paradigms: reconstruction-oriented VAEs and semantic-oriented vision encoders. To represent the former, we select the standard VAE \cite{kingma2013auto} and VA-VAE \cite{yao2025reconstruction}. For the latter, we evaluate MAE \cite{he2022masked}, DINOv2 \cite{oquab2023dinov2}, and SigLIP2 \cite{tschannen2025siglip}. When evaluating the semantic vision encoders, we adopt the methodology established by RAE \cite{zheng2025diffusion}, wherein the encoder remains frozen and only the decoder is trained to reconstruct the input images. Across these five latent representations, we comprehensively evaluate both reconstruction fidelity and generative performance within the continuous autoregressive modeling paradigm.

We assess reconstruction quality using four standard metrics: rFID, PSNR, LPIPS, and SSIM. Here, rFID denotes the FID score \cite{heusel2017gans} computed on the reconstructed ImageNet \cite{russakovsky2015imagenet} validation set. Let $f$ denote the spatial compression ratio and $d$ represent the channel dimension of the latent representations, which jointly decide the information capacity of latents. As detailed in Tab. \ref{tab:Reconstruction}, semantic-oriented representations possess a significantly larger information capacity compared to their reconstruction-oriented counterparts, owing to their higher channel dimensions. In terms of distribution reconstruction (rFID), both types of representations achieve comparable results. However, from the perspective of pixel-wise fidelity, semantic-oriented representations are generally inferior, exhibiting substantially lower PSNR values. The notable exception is MAE, which achieves the highest PSNR overall. This is likely attributable to its Masked Image Modeling (MIM) pretraining objective, which is inherently designed for pixel-level reconstruction.

Following the reconstruction analysis, we investigate the continuous autoregressive generation performance of these representations. As presented in Tab. \ref{tab:Generation}, we draw two primary conclusions:
\textbf{First}, the direct integration of semantic-oriented representations yields consistently and substantially poorer generative performance compared to reconstruction-oriented VAEs. For instance, SigLIP2 and MAE achieve FID scores of 34.084 and 67.642, respectively. This demonstrates that comparable reconstruction fidelity does not guarantee equivalent generative capability. This finding aligns with observations reported in RAE \cite{zheng2025diffusion}.
\textbf{Second}, among the three evaluated semantic encoders, DINOv2 delivers the best generative performance, while MAE performs the worst. This hierarchy of semantic encoder efficacy is consistent with the conclusions drawn in RAE \cite{zheng2025diffusion}.

\begin{table}[t]
    \setlength{\tabcolsep}{10pt}
    \centering
    \caption{Generation performance of the five generative latents. The direct combination of representation autoencoders and AR model leads to poor generative performance.}
    \label{tab:Generation}
    \begin{tabular}{c|ccccc}
    \toprule
    Generation  &  VAE  &  VA-VAE  &  DINOv2  &  SigLIP2  &  MAE  \\
    \midrule[0.1em]
    Metric & f16d16 & f16d32  &  f16d768  &  f16d768  &  f16d768  \\
    \midrule
    gFID$\downarrow$ & 3.237  & 5.983  & 15.137  & 34.084  & 67.642  \\
    IS$\uparrow$   &259.168   &206.839   &108.384   &46.711   &24.067   \\
    Precision$\uparrow$  & 0.135  & 0.097  & 0.223  & 0.218  & 0.085  \\
    Recall$\uparrow$   & 0.716  & 0.779  & 0.639  & 0.664  & 0.914  \\
    \bottomrule
    \end{tabular}
\end{table}

\section{Method}

\subsection{Verification of Designs in Diffusion Models}
RAE \cite{zheng2025diffusion, tong2026scaling, tong2026beyond} reveals that model width and noise schedule play a key role. 
Given that the continuous autoregressive model also has a diffusion denoiser MLP for token distribution modeling, we first verify the effectiveness of these two designs. Existing continuous autoregressive methods \cite{li2024autoregressive, yu2025frequency} conclude that larger models usually leads to better performance. We thus adopt model width \textit{w}=1280 by default, and ablate the effectiveness of the noise schedule. As shown in Tab. \ref{tab:schedule} in the experimental part, dimension-dependent schedule \cite{zheng2025diffusion} doesn't achieve consistent improvements across the three representation autoencoders. Consequently, we adopt the usual noise schedule in the following experiments.

Apart from the above verification, considering the unique properties of continuous autoregressive models, we identify two crucial factors that affect the performance of the high-dimensional semantic-oriented representations: 1) token Variance (complex token-wise distribution modeling); 2) exposure bias (the high-dimensionality amplified training-inference gap).

\subsection{Token Variance}  \label{sec:token_variance}

We first investigate the impact of token variance. Unlike the image-level distribution modeling typical of diffusion frameworks, autoregressive (AR) models capture token-wise distributions sequentially. Because each predicted token is conditioned on preceding ones, a larger variance among tokens can significantly exacerbate modeling and training difficulty. To investigate this hypothesis, we compute the statistics of the token distribution by calculating the mean and variance of each token along the channel dimension. This process yields a spatial mean map (\textit{mean-map}) and variance map (\textit{var-map}) for each image.

\begin{figure}[h]
    \begin{center}
	\includegraphics[width=0.98\linewidth]{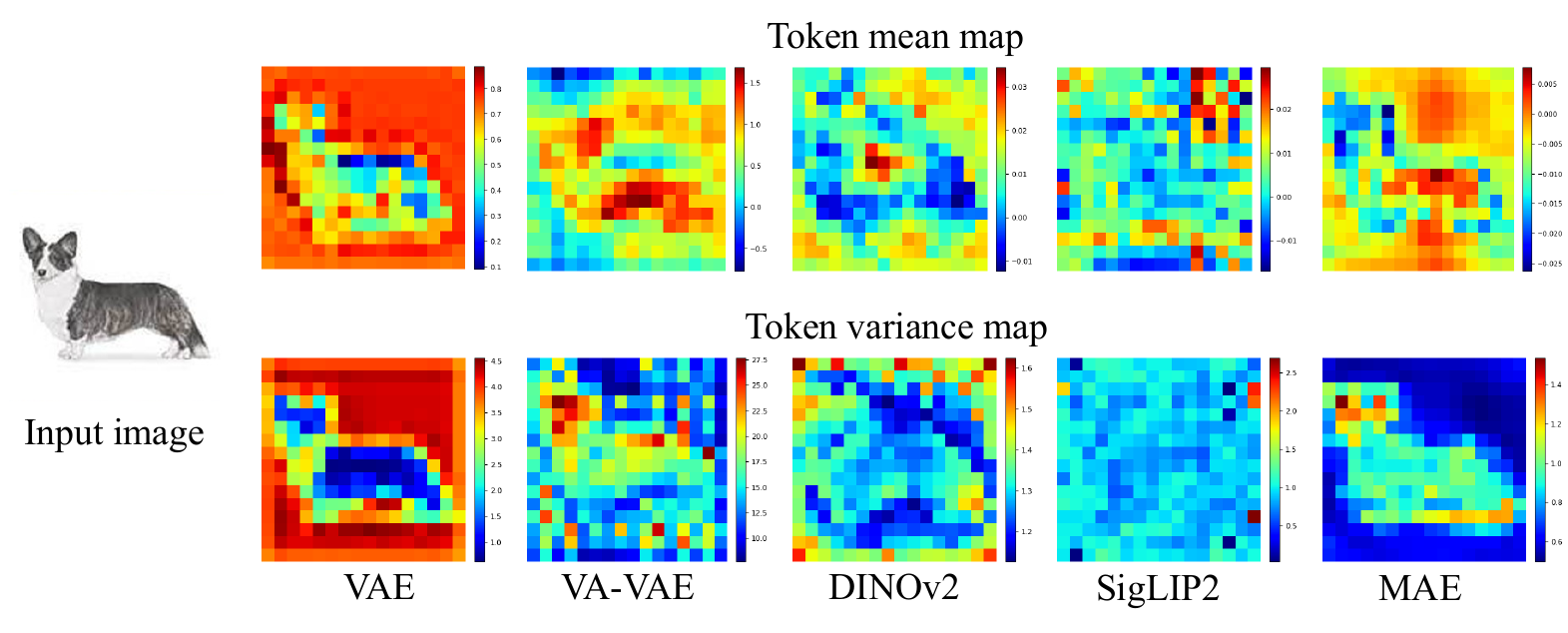}
    \end{center}
    \setlength{\abovecaptionskip}{-0.2cm}
    \setlength{\belowcaptionskip}{-0.2cm}
    \caption{The mean and variance map visualization of the five generative latents. Token-wise distribution variance obviously exists and may influence the modeling difficulty of AR models, as verified by the poor generation performance shown in Tab. \ref{tab:Generation}. }
    \label{fig:mean_variance}
    \vspace{-3mm}
\end{figure}

As illustrated in Fig. \ref{fig:mean_variance}, we visualize the \textit{mean-map} and \textit{var-map} across five representations: reconstruction-oriented latents (VAE and VA-VAE) and semantic-oriented latents (DINOv2, SigLIP2, and MAE). The inter-token variance is visually evident, and structural patterns are clearly distinguishable. To quantify this, we measure the spatial diversity of the token variance map, which we formally denote as \textit{token variance}. As detailed in Tab. \ref{tab:mean_variance}, we extract three key statistics: (a) the spatial mean of the \textit{mean-map}; (b) the spatial mean of the \textit{var-map}; and (c) the spatial variance of the \textit{var-map}, denoted as \textit{token variance}.

\begin{table}[t]
    \setlength{\tabcolsep}{10pt}
    \centering
    \caption{Different statistics of tokens from the generative latents. Statistic (a) represents the spatial mean of the mean-map. Statistic (b) represents the spatial mean of the var-map. Statistic (c) represents the spatial variance of the var-map (token variance). Lower token variance usually leads to better generation performance.
    }
    \label{tab:mean_variance}
    \begin{tabular}{l|ccccc}
    \toprule
    Statistic  &  VAE  &  VA-VAE  &  DINOv2  &  SigLIP2  &  MAE  \\
    \midrule[0.1em]
    Statistic (a)   & 0.1889 & 0.1853  & 0.0113 & 0.0001 & -0.0025  \\
    Statistic (b)   & 1.2966 & 13.1932 & 1.3014 & 1.0072 & 0.9503  \\
    Statistic (c)   & 0.6158 & 18.4432 & 0.0100 & 0.0968 & 0.0858  \\
    \bottomrule
    \end{tabular}
\end{table}

Our analysis reveals that among reconstruction-oriented representations, the \textit{token variance} of the standard VAE is significantly lower than that of VA-VAE, which corresponds to its superior generative performance. Similarly, among semantic-oriented representations, DINOv2 exhibits a significantly lower \textit{token variance} compared to both SigLIP2 and MAE, which also corresponds to its better generation quality. These observations align with our initial assumption regarding token modeling difficulty.

\begin{figure}[t]
    \begin{center}
	\includegraphics[width=0.98\linewidth]{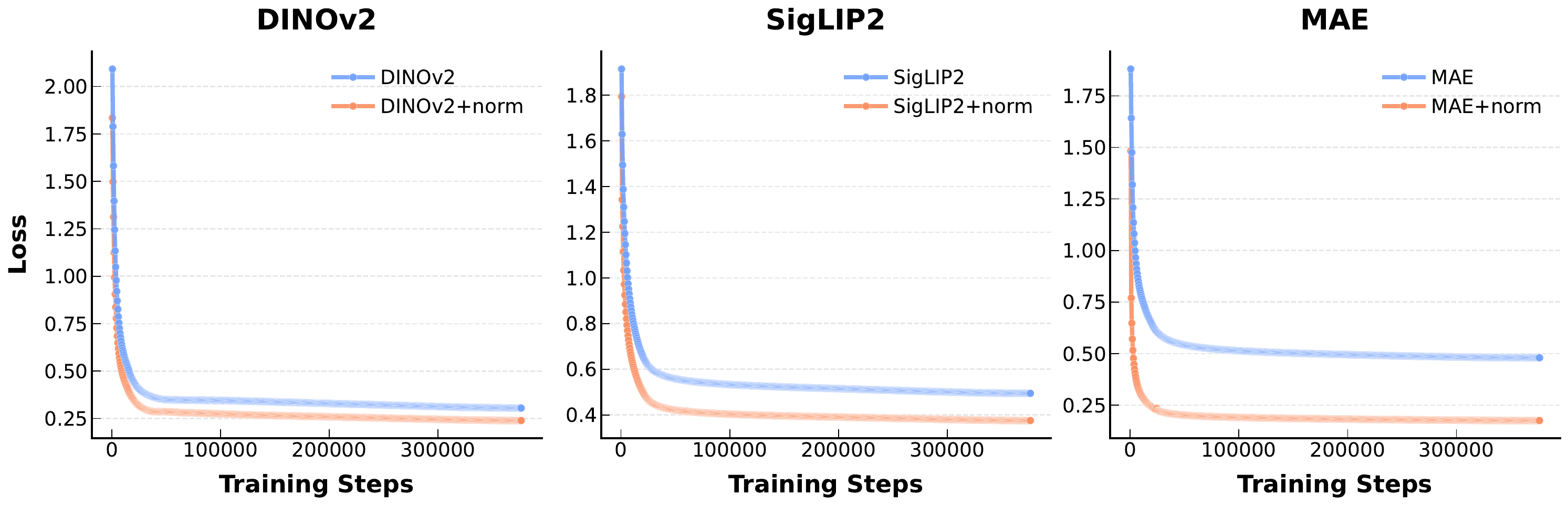}
    \end{center}
    \setlength{\abovecaptionskip}{-0.2cm}
    \setlength{\belowcaptionskip}{-0.2cm}
    \caption{The training loss curve comparison on the three representation autoencoders. Token normalization introduces lower training loss and fast convergence.}
    \label{fig:training_loss}
    \vspace{-3mm}
\end{figure}

Building on these findings, we propose the token normalization strategy, normalizing the distribution of each token to enforce a zero mean and unit variance ($\mu=0, \sigma^2=1$). By minimizing the variance between tokens, we expect a corresponding reduction in the AR modeling burden. Empirical evaluations confirm this hypothesis: training on normalized tokens yields a consistently lower training loss and substantially accelerates convergence. As demonstrated by the training curves in Fig. \ref{fig:training_loss}, while the initial losses are comparable, the normalized approach rapidly descends to a lower loss plateau.  Crucially, this normalization strategy achieves faster convergence across all three evaluated representation autoencoders, underscoring its generalizability and overall effectiveness.

\subsection{Exposure Bias}

It is important to note that while token normalization successfully reduces modeling difficulty and accelerates convergence, it does not inherently guarantee improved inference performance. This discrepancy stems from the inherent exposure bias prevalent in autoregressive models. During the training phase, predictions are conditioned on ground-truth tokens (teacher forcing). Conversely, during inference, subsequent predictions rely entirely on previously generated tokens. Consequently, if a generated token deviates even slightly from the target distribution, this error propagates and compounds across subsequent steps. This training-inference gap severely degrades generation quality, explaining why token normalization alone is insufficient. Furthermore, because the channel dimension influences the modeling capacity and difficulty in autoregressive frameworks, this exposure bias is significantly amplified when employing high-dimensional, semantic-oriented representations. This is also verified with the poor image quality in the left part of Fig. \ref{fig:visual_comparison}.

To this end, we propose enhancing inference robustness by introducing Gaussian noise perturbation during the training stage. Specifically, by adding Gaussian noise to each token, we force the model to condition its predictions on perturbed representations rather than ground-truth tokens. Consequently, the AR model learns to correct for deviations and becomes significantly more tolerant to the distributional shifts of predicted tokens during the inference stage. 

Beyond the direct gains achieved by noise perturbation in isolation, we observe a synergistic performance elevation when combining it with token normalization. Token normalization effectively addresses the token variance issue, easing the overall modeling difficulty during training. Concurrently, noise perturbation mitigates the training-inference gap, directly enhancing inference fidelity. Ultimately, the conjunction of these two strategies simultaneously optimizes both the training and inference phases, yielding substantially better generative performance than either technique applied independently. 

\section{Experiments}

\subsection{Experiment Setups}

\noindent {\bf Training details.}
We adopt the continuous autoregressive model as backbone, coupled with the latent space from both the three representation autoencoders and two reconstruction-oriented autoencoders (VAE and VA-VAE). The autoregressive part causally models each token in raster-scan order. Then, the token-wise output of the autoregressive part independently serves as the condition of the diffusion MLP.  
We train the model for 300 epochs. The model width is set to 1280. Additional implementation details are available in the supplementary material.

\noindent {\bf Evaluation metrics.}
For image reconstruction, we adopt reconstruction FID (rFID) \cite{heusel2017gans}, PSNR, LPIPS \cite{zhang2018unreasonable}, and SSIM \cite{wang2004image}. For image generation, we report generation FID (gFID) \cite{heusel2017gans}, Inception Score (IS) \cite{salimans2016improved}, Perception, and Recall.

\begin{figure}[ht]
    \begin{center}
	\includegraphics[width=0.98\linewidth]{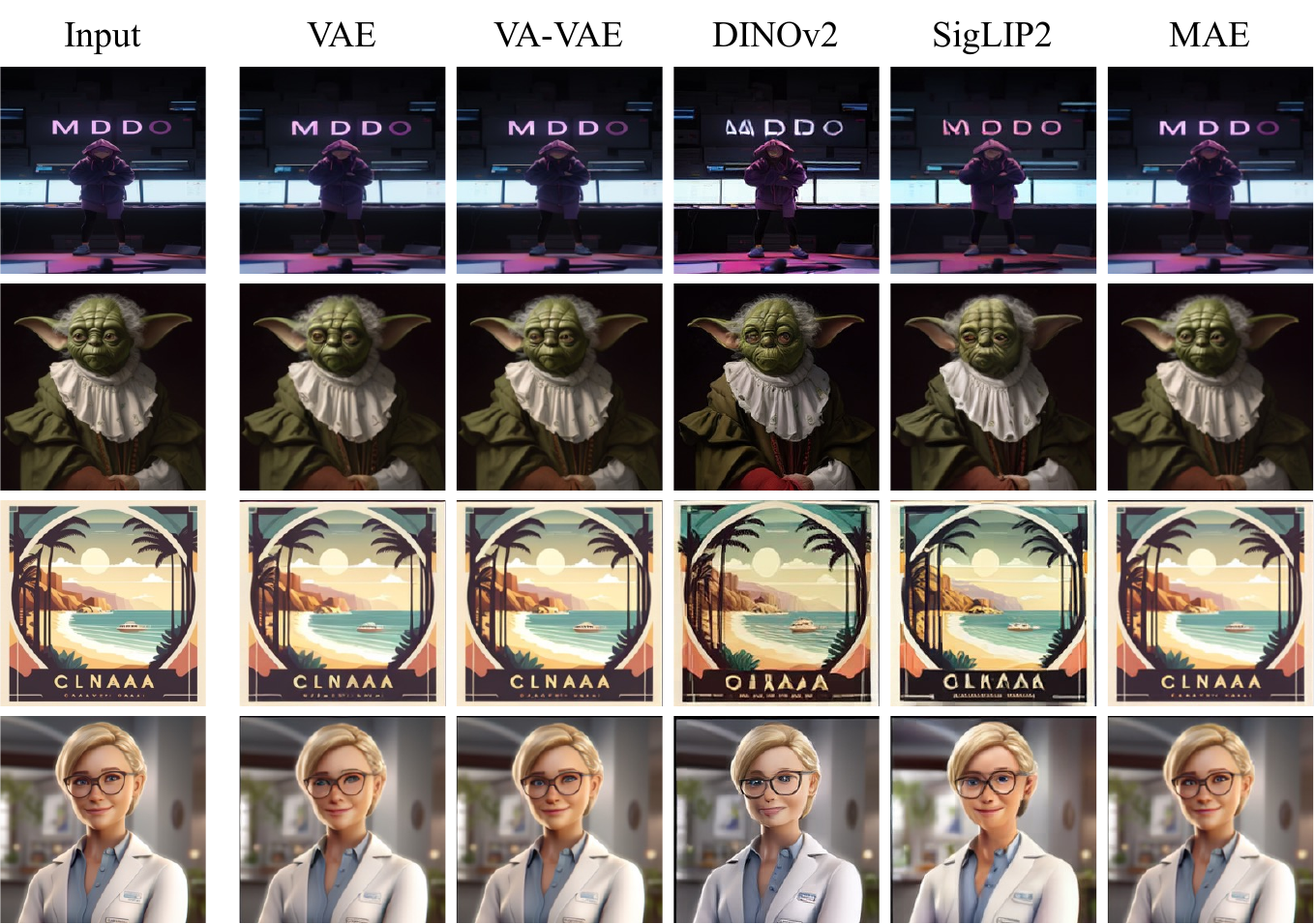}
    \end{center}
    \setlength{\abovecaptionskip}{-0.2cm}
    \setlength{\belowcaptionskip}{-0.2cm}
    \caption{The reconstruction results comparison between VAE, VA-VAE, and three representation autoencoders. VAE type usually performs better in fidelity and details.}
    \label{fig:reconstruction}
    \vspace{-3mm}
\end{figure}

\subsection{Reconstruction Results}

\noindent {\bf Quantitative comparison.}
We present the reconstruction performance comparison of various diffusion latents in Tab. \ref{tab:Reconstruction}. Similar to the conclusion in RAE, we also reveal that the representation autoencoder bears comparable rFID performance than VAE types. However, the fidelity of representation autoencoder, demonstrated by PSNR and SSIM, is inferior to VAEs. The only exception is MAE, which may be relevant to its reconstruction target.

\noindent {\bf Visual comparison.}
To visualize the reconstruction result, we present the visual comparison in Fig. \ref{fig:reconstruction}. The representation autoencoder is able to reconstruct most of the information, including structure, texture, and color. However, the details are relatively poor compared to the VAEs. Concretely, the words in the first and third rows are failed to be perfectly reconstructed. The faces in the second and last rows are blurry and poorly structured. Besides, we find that the reconstruction of DINOv2 is prone to have severe color shift. 

Given that the representation autoencoder has higher information capacity due to significantly larger channel dimensions, the weak detail fidelity suggests that it still has room for improvement in detailed information.

\begin{table}[ht]
    \setlength{\tabcolsep}{11pt}
    \centering
    \caption{The generation performance of the continuous autoregressive model on both VAEs, three representative representations autoencoders, and our improved versions.}
    \label{tab:performance}
    \begin{tabular}{l|rrcc}
    \toprule
    Metrics  &  gFID  &  IS  &  Perception  &  Recall   \\
    \midrule
    VAE                   & 3.237   &259.168   & 0.741 & 0.649  \\
    VA-VAE                & 5.983   &206.839   & 0.679 & 0.654  \\
    \midrule
    DINOv2 (Baseline)     & 15.137  &108.384   &0.702  &0.474    \\
    +Norm                 & 18.090  &99.342   &0.682  &0.484   \\
    +Noise                & 9.516   &144.619   &0.774  &0.429   \\
    RAE-AR (+Norm+Noise)  & 7.494   &165.588   &0.804  &0.405   \\
    \midrule
    SigLIP2 (Baseline)     & 34.084  &46.711   &0.551  &0.518    \\
    +Norm                  & 35.703  &45.524   &0.601  &0.508   \\
    +Noise                 & 6.767   &193.059   &0.847  &0.330   \\
    RAE-AR (+Norm+Noise)   & 6.091   &241.768   &0.869  &0.335   \\
    \midrule
    MAE (Baseline)         & 67.642  &24.067   &0.343  &0.582    \\
    +Norm                  & 54.053  &28.780   &0.428  &0.570   \\
    +Noise                 & 22.595   &73.463   &0.638  &0.439   \\
    RAE-AR (+Norm+Noise)   & 9.083   &129.770   &0.833  &0.290   \\
    \bottomrule
    \end{tabular}
\end{table}

\subsection{Generation Results}

\noindent {\bf Quantitative comparison.}
In Tab. \ref{tab:performance}, we present the generative performance of the continuous autoregressive model on 1) the two reconstruction-oriented autoencoders, including VAE and VA-VAE; 2) the three representation autoencoders, including DINOv2, SigLIP2, and MAE; 3) our proposed strategies on these three representation autoencoders, including token normalization and noise perturbation.

We achieve several conclusions from the results. (1) Although representation autoencoders has comparable reconstruction performance as the reconstruction-oriented autoencoders, the autoregressive generative performance on these representation autoencoders are terrible. (2) Among these three representation autoencoders, DINOv2 has the best performance and MAE performs the worst. This observation is consistent with the conclusion in RAE. (3) The exposure bias problem directly influence the performance and is amplified with the high-dimension representation autoencoders. Therefore, the noise perturbation operation mitigates such bias and contributes considerably to the performance. (4) The token normalization operation, as explained in Sec. \ref{sec:token_variance}, effectively eases the training difficulty. While, normalization-only doesn't address the training-inference gap, and may not lead to better performance. (5) The combination of token normalization and noise perturbation contributes the the largest performance improvement, validating the necessity and effectiveness of the token normalization operation.

\begin{figure}[t]
    \begin{center}
        \includegraphics[width=0.98\linewidth]{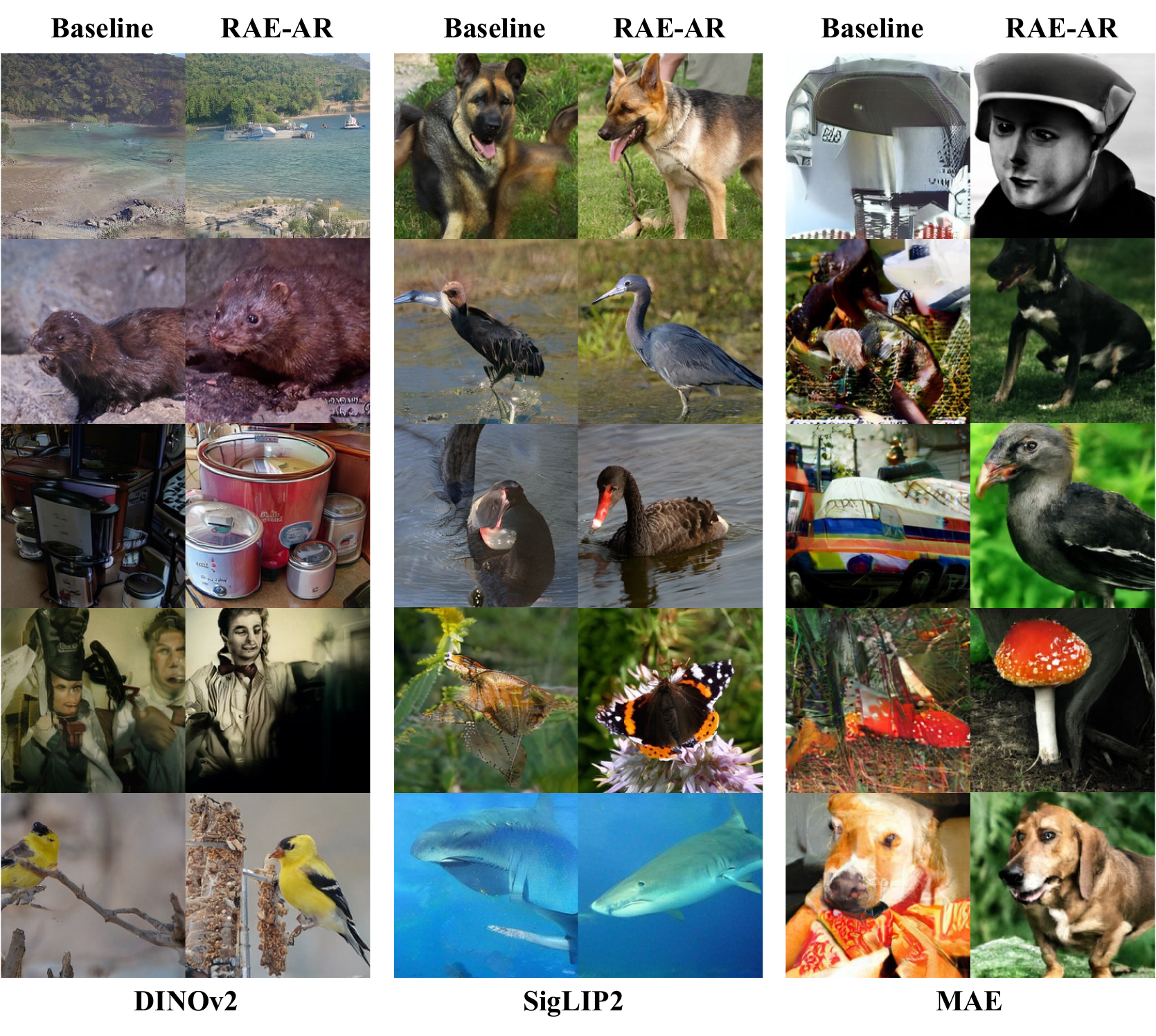}
    \end{center}
    \setlength{\abovecaptionskip}{-0.2cm}
    \setlength{\belowcaptionskip}{-0.2cm}
    \caption{The visual generation comparison between the baseline and our RAE-AR on the three representation autoencoders.}
    \label{fig:visual_comparison}
    \vspace{-3mm}
\end{figure}

\noindent {\bf Visual comparison.}
As illustrated in Fig. \ref{fig:visual_comparison}, we provide a qualitative comparison between the baseline and our proposed RAE-AR across three representation autoencoders. The baseline results are characterized by distorted structures, blurred textures, and a lack of fine detail. Notably, complex semantic structures, such as human faces, fail to be faithfully reconstructed. In contrast, the enhanced RAE-AR consistently generates images with superior structural integrity and more realistic details.


\subsection{More Ablations}

Apart from the reconstruction and generation results above, we also conduct three important ablations: 1) the verification of dimension-dependent schedule; 2) the correlation between noise level and generative performance; 3) the verification of the proposed strategies on the mask-based autoregressive model.

\noindent {\bf Dimension-dependent noise schedule.}
In Tab. \ref{tab:schedule}, we present the ablation results on the dimension-dependent noise schedule. On the native representation autoencoders (the first two rows), this schedule design is better for DINOv2 and MAE, but degenerates for SigLIP2 (34.084 FID to 38.286). Based on our improved version RAE-AR (the last two rows), this schedule also fail to achieve consistent improvement, with worse performance on SigLIP2 (6.091 FID to 6.431) and MAE.

\begin{table}[ht]
    \setlength{\tabcolsep}{12pt}
    \centering
    \caption{The effectiveness verification of the dimension-dependent noise schedule design on continuous autoregressive model.}
    \label{tab:schedule}
    \begin{tabular}{l|cccc}
    \toprule
    Metric      &  gFID   &  IS      &  Perception  &  Recall   \\
    \midrule
    DINOv2      & 15.137  & 108.384  & 0.702   & 0.474     \\
    DINOv2 (Schedule)  & 13.734    & 113.372   & 0.718     & 0.478    \\
    RAE-AR      & 7.494   & 165.588  & 0.804   &0.405   \\
    RAE-AR (Schedule)  & 6.597     & 185.945   & 0.834     & 0.378    \\
    \midrule
    SigLIP2     & 34.084  & 46.711   & 0.551   & 0.518     \\
    SigLIP2 (Schedule)  & 38.286   & 41.082    & 0.597    & 0.510      \\
    RAE-AR   & 6.091   & 241.768   & 0.869     & 0.335   \\
    RAE-AR (Schedule)   & 6.431    & 248.423   & 0.888     & 0.298    \\
    \midrule
    MAE         & 67.642  & 24.067   & 0.343        & 0.582     \\
    MAE (Schedule)  & 47.531  & 32.610   & 0.438        & 0.587     \\
    RAE-AR    & 9.083   &129.770   &0.833  &0.290   \\
    RAE-AR (Schedule)   & 10.393    & 202.313   & 0.796     & 0.271    \\
    \bottomrule
    \end{tabular}
\end{table}

\noindent {\bf The noise level and performance.}
In Tab. \ref{tab:performance}, we validate the effectiveness of the noise perturbation. In this section, we further discuss the optimality of the noise level selection. 
Specifically, given an input sequence mapped to its continuous embedding representation $E \in \mathbb{R}^{N \times d}$, where $N$ is the sequence length and $d$ is the hidden dimension, we apply an additive perturbation matrix $\Delta$ directly to the token embeddings. The noise-injected representations, denoted as $\tilde{E}$, are formulated as:
\begin{equation}
    \tilde{E} = E + \alpha \cdot \Delta,
\end{equation}
where $\Delta \sim \mathcal{N}(0, \mathbf{I})$ is a noise matrix sampled from a standard Gaussian distribution, and $\alpha \in \mathbb{R}^{+}$ is the scaling hyperparameter that determines the intensity of the injected noise. 

\begin{figure}[h]
    \begin{center}
        \includegraphics[width=0.98\linewidth]{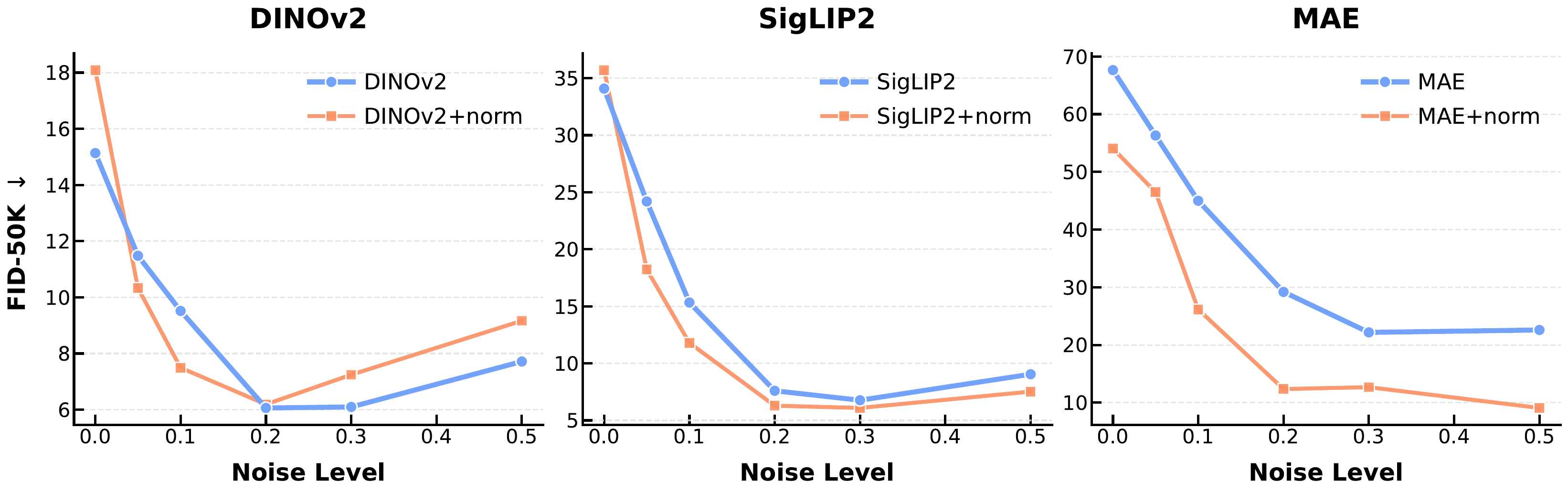}
    \end{center}
    \setlength{\abovecaptionskip}{-0.2cm}
    \setlength{\belowcaptionskip}{-0.2cm}
    \caption{The relationship between noise perturbation level and generation performance on the three representation autoencoders.}
    \label{fig:noise_level}
    \vspace{-3mm}
\end{figure}

As shown in Fig. \ref{fig:noise_level}, we ablate different noise level that are injected into the tokens, and systematically vary $\alpha$ over a predefined set of values (e.g., $\alpha \in \{0.0, 0.05, 0.1, 0.2, 0.3, 0.5\}$) to observe its impact on the model's training dynamics and final performance. As the performance curve illustrates, the performance with noise level follows a similar trend of first improves and then degrades. Thus, these representation autoencoders respectively have a moderate noise level that achieves the optimal performance. 
However, when $\alpha$ surpasses an optimal threshold, the magnitude of the perturbation overwhelms the original semantic signal encapsulated within the tokens, inevitably leading to performance decline. For example, SigLIP2 starts to degrade when the noise level $\alpha$ is larger than 0.3.

\noindent {\bf Application to mask-based autoregressive model.}
In the main part, our experiments are conducted on the autoregressive models with causal raster-scan order. Another mainstream regression paradigm employs masked and parallelized regression \cite{li2024autoregressive}, which predicts the masked tokens given the unmasked ones. 
In this section, we verify the effectiveness of the proposed strategies on these mask-based generation. As shown in Tab. \ref{tab:MAR}, we experiment with the SigLIP2 representation and the noise level of 0.1. The noise injection brings substantial performance lift. The combination of token normalization and noise injection operations leads to the best performance. This conclusion is consistent with that on the causal autoregressive model, demonstrating the generality and robustness of the proposed method.

\begin{table}[ht]
    \setlength{\tabcolsep}{12pt}
    \centering
    \caption{The generation performance of the SigLIP2 latent representation on mask-based autoregressive model.}
    \label{tab:MAR}
    \begin{tabular}{l|cccc}
    \toprule
    Metirc  &  gFID  &  IS  &  Perception  &  Recall    \\
    \midrule
    SigLIP2 (Baseline)      & 12.624 &124.423   & 0.727  & 0.478  \\
    +Norm                   & 12.857 &117.808   & 0.735  & 0.475 \\
    +Noise                  & 9.662  &149.770   & 0.762  & 0.441  \\
    RAE-AR (+Norm+Noise)    & 8.957  &152.389   & 0.782  & 0.440  \\
    \bottomrule
    \end{tabular}
\end{table}

\section{Conclusion}
This paper challenges the long-standing reliance on VAE-based encoders in generative modeling by exploring the potential of representation-based autoencoders within continuous autoregressive (AR) frameworks. While representation encoders have historically been viewed as unsuitable for generation, we identify that their integration into AR models (termed RAE-AR) is primarily hindered by two factors: the difficulty of modeling complex token-wise distributions and an intensified exposure bias resulting from high-dimensional latent spaces. To overcome these obstacles, we proposes a dual-strategy approach: implementing distribution normalization to simplify token modeling and utilizing Gaussian noise injection during the training phase to bolster inference robustness. Experimental results suggest that these refinements allow representation autoencoders to match the performance of VAEs, offering a viable path toward unifying visual representation learning.

\clearpage  


%
%
\bibliographystyle{splncs04}
\bibliography{main}

@article{zheng2025diffusion,
  title={Diffusion transformers with representation autoencoders},
  author={Zheng, Boyang and Ma, Nanye and Tong, Shengbang and Xie, Saining},
  journal={arXiv preprint arXiv:2510.11690},
  year={2025}
}

@article{tong2026scaling,
  title={Scaling Text-to-Image Diffusion Transformers with Representation Autoencoders},
  author={Tong, Shengbang and Zheng, Boyang and Wang, Ziteng and Tang, Bingda and Ma, Nanye and Brown, Ellis and Yang, Jihan and Fergus, Rob and LeCun, Yann and Xie, Saining},
  journal={arXiv preprint arXiv:2601.16208},
  year={2026}
}

@article{shi2025latent,
  title={Latent diffusion model without variational autoencoder},
  author={Shi, Minglei and Wang, Haolin and Zheng, Wenzhao and Yuan, Ziyang and Wu, Xiaoshi and Wang, Xintao and Wan, Pengfei and Zhou, Jie and Lu, Jiwen},
  journal={arXiv preprint arXiv:2510.15301},
  year={2025}
}

@article{shi2025svg,
  title={SVG-T2I: Scaling Up Text-to-Image Latent Diffusion Model Without Variational Autoencoder},
  author={Shi, Minglei and Wang, Haolin and Zhang, Borui and Zheng, Wenzhao and Zeng, Bohan and Yuan, Ziyang and Wu, Xiaoshi and Zhang, Yuanxing and Yang, Huan and Wang, Xintao and others},
  journal={arXiv preprint arXiv:2512.11749},
  year={2025}
}

@article{yue2025uniflow,
  title={UniFlow: A Unified Pixel Flow Tokenizer for Visual Understanding and Generation},
  author={Yue, Zhengrong and Zhang, Haiyu and Zeng, Xiangyu and Chen, Boyu and Wang, Chenting and Zhuang, Shaobin and Dong, Lu and Du, KunPeng and Wang, Yi and Wang, Limin and others},
  journal={arXiv preprint arXiv:2510.10575},
  year={2025}
}

@article{chen2025aligning,
  title={Aligning visual foundation encoders to tokenizers for diffusion models},
  author={Chen, Bowei and Bi, Sai and Tan, Hao and Zhang, He and Zhang, Tianyuan and Li, Zhengqi and Xiong, Yuanjun and Zhang, Jianming and Zhang, Kai},
  journal={arXiv preprint arXiv:2509.25162},
  year={2025}
}

@inproceedings{yao2025reconstruction,
  title={Reconstruction vs. generation: Taming optimization dilemma in latent diffusion models},
  author={Yao, Jingfeng and Yang, Bin and Wang, Xinggang},
  booktitle={Proceedings of the Computer Vision and Pattern Recognition Conference},
  pages={15703--15712},
  year={2025}
}

@article{yu2024representation,
  title={Representation alignment for generation: Training diffusion transformers is easier than you think},
  author={Yu, Sihyun and Kwak, Sangkyung and Jang, Huiwon and Jeong, Jongheon and Huang, Jonathan and Shin, Jinwoo and Xie, Saining},
  journal={arXiv preprint arXiv:2410.06940},
  year={2024}
}

@article{oquab2023dinov2,
  title={Dinov2: Learning robust visual features without supervision},
  author={Oquab, Maxime and Darcet, Timoth{\'e}e and Moutakanni, Th{\'e}o and Vo, Huy and Szafraniec, Marc and Khalidov, Vasil and Fernandez, Pierre and Haziza, Daniel and Massa, Francisco and El-Nouby, Alaaeldin and others},
  journal={arXiv preprint arXiv:2304.07193},
  year={2023}
}

@article{simeoni2025dinov3,
  title={Dinov3},
  author={Sim{\'e}oni, Oriane and Vo, Huy V and Seitzer, Maximilian and Baldassarre, Federico and Oquab, Maxime and Jose, Cijo and Khalidov, Vasil and Szafraniec, Marc and Yi, Seungeun and Ramamonjisoa, Micha{\"e}l and others},
  journal={arXiv preprint arXiv:2508.10104},
  year={2025}
}

@article{tschannen2025siglip,
  title={Siglip 2: Multilingual vision-language encoders with improved semantic understanding, localization, and dense features},
  author={Tschannen, Michael and Gritsenko, Alexey and Wang, Xiao and Naeem, Muhammad Ferjad and Alabdulmohsin, Ibrahim and Parthasarathy, Nikhil and Evans, Talfan and Beyer, Lucas and Xia, Ye and Mustafa, Basil and others},
  journal={arXiv preprint arXiv:2502.14786},
  year={2025}
}

@inproceedings{assran2023self,
  title={Self-supervised learning from images with a joint-embedding predictive architecture},
  author={Assran, Mahmoud and Duval, Quentin and Misra, Ishan and Bojanowski, Piotr and Vincent, Pascal and Rabbat, Michael and LeCun, Yann and Ballas, Nicolas},
  booktitle={Proceedings of the IEEE/CVF conference on computer vision and pattern recognition},
  pages={15619--15629},
  year={2023}
}

@article{kingma2013auto,
  title={Auto-encoding variational bayes},
  author={Kingma, Diederik P and Welling, Max},
  journal={arXiv preprint arXiv:1312.6114},
  year={2013}
}

@inproceedings{he2022masked,
  title={Masked autoencoders are scalable vision learners},
  author={He, Kaiming and Chen, Xinlei and Xie, Saining and Li, Yanghao and Doll{\'a}r, Piotr and Girshick, Ross},
  booktitle={Proceedings of the IEEE/CVF conference on computer vision and pattern recognition},
  pages={16000--16009},
  year={2022}
}

@inproceedings{radford2021learning,
  title={Learning transferable visual models from natural language supervision},
  author={Radford, Alec and Kim, Jong Wook and Hallacy, Chris and Ramesh, Aditya and Goh, Gabriel and Agarwal, Sandhini and Sastry, Girish and Askell, Amanda and Mishkin, Pamela and Clark, Jack and others},
  booktitle={International conference on machine learning},
  pages={8748--8763},
  year={2021},
  organization={PmLR}
}

@article{yu2025videomar,
  title={Videomar: Autoregressive video generatio with continuous tokens},
  author={Yu, Hu and Gong, Biao and Yuan, Hangjie and Zheng, DanDan and Chai, Weilong and Chen, Jingdong and Zheng, Kecheng and Zhao, Feng},
  journal={arXiv preprint arXiv:2506.14168},
  year={2025}
}

@article{yu2025frequency,
  title={Frequency autoregressive image generation with continuous tokens},
  author={Yu, Hu and Luo, Hao and Yuan, Hangjie and Rong, Yu and Zhao, Feng},
  journal={arXiv preprint arXiv:2503.05305},
  year={2025}
}

@article{agarwal2025cosmos,
  title={Cosmos world foundation model platform for physical ai},
  author={Agarwal, Niket and Ali, Arslan and Bala, Maciej and Balaji, Yogesh and Barker, Erik and Cai, Tiffany and Chattopadhyay, Prithvijit and Chen, Yongxin and Cui, Yin and Ding, Yifan and others},
  journal={arXiv preprint arXiv:2501.03575},
  year={2025}
}

@article{tian2024visual,
  title={Visual autoregressive modeling: Scalable image generation via next-scale prediction},
  author={Tian, Keyu and Jiang, Yi and Yuan, Zehuan and Peng, Bingyue and Wang, Liwei},
  journal={Advances in neural information processing systems},
  volume={37},
  pages={84839--84865},
  year={2024}
}

@article{li2024autoregressive,
  title={Autoregressive image generation without vector quantization},
  author={Li, Tianhong and Tian, Yonglong and Li, He and Deng, Mingyang and He, Kaiming},
  journal={Advances in Neural Information Processing Systems},
  volume={37},
  pages={56424--56445},
  year={2024}
}

@article{fan2024fluid,
  title={Fluid: Scaling autoregressive text-to-image generative models with continuous tokens},
  author={Fan, Lijie and Li, Tianhong and Qin, Siyang and Li, Yuanzhen and Sun, Chen and Rubinstein, Michael and Sun, Deqing and He, Kaiming and Tian, Yonglong},
  journal={arXiv preprint arXiv:2410.13863},
  year={2024}
}

@inproceedings{chang2022maskgit,
  title={Maskgit: Masked generative image transformer},
  author={Chang, Huiwen and Zhang, Han and Jiang, Lu and Liu, Ce and Freeman, William T},
  booktitle={Proceedings of the IEEE/CVF conference on computer vision and pattern recognition},
  pages={11315--11325},
  year={2022}
}

@inproceedings{esser2021taming,
  title={Taming transformers for high-resolution image synthesis},
  author={Esser, Patrick and Rombach, Robin and Ommer, Bjorn},
  booktitle={Proceedings of the IEEE/CVF conference on computer vision and pattern recognition},
  pages={12873--12883},
  year={2021}
}

@article{sun2024autoregressive,
  title={Autoregressive model beats diffusion: Llama for scalable image generation},
  author={Sun, Peize and Jiang, Yi and Chen, Shoufa and Zhang, Shilong and Peng, Bingyue and Luo, Ping and Yuan, Zehuan},
  journal={arXiv preprint arXiv:2406.06525},
  year={2024}
}

@article{brown2020language,
  title={Language models are few-shot learners},
  author={Brown, Tom B},
  journal={arXiv preprint arXiv:2005.14165},
  year={2020}
}

@inproceedings{lee2022autoregressive,
  title={Autoregressive image generation using residual quantization},
  author={Lee, Doyup and Kim, Chiheon and Kim, Saehoon and Cho, Minsu and Han, Wook-Shin},
  booktitle={Proceedings of the IEEE/CVF Conference on Computer Vision and Pattern Recognition},
  pages={11523--11532},
  year={2022}
}

@article{devlin2018bert,
  title={Bert: Pre-training of deep bidirectional transformers for language understanding},
  author={Devlin, Jacob},
  journal={arXiv preprint arXiv:1810.04805},
  year={2018}
}

@inproceedings{rombach2022high,
  title={High-resolution image synthesis with latent diffusion models},
  author={Rombach, Robin and Blattmann, Andreas and Lorenz, Dominik and Esser, Patrick and Ommer, Bj{\"o}rn},
  booktitle={Proceedings of the IEEE/CVF conference on computer vision and pattern recognition},
  pages={10684--10695},
  year={2022}
}

@inproceedings{chen2025masked,
  title={Masked autoencoders are effective tokenizers for diffusion models},
  author={Chen, Hao and Han, Yujin and Chen, Fangyi and Li, Xiang and Wang, Yidong and Wang, Jindong and Wang, Ze and Liu, Zicheng and Zou, Difan and Raj, Bhiksha},
  booktitle={Forty-second International Conference on Machine Learning},
  year={2025}
}

@inproceedings{chen2025dc,
  title={Dc-ae 1.5: Accelerating diffusion model convergence with structured latent space},
  author={Chen, Junyu and Zou, Dongyun and He, Wenkun and Chen, Junsong and Xie, Enze and Han, Song and Cai, Han},
  booktitle={Proceedings of the IEEE/CVF International Conference on Computer Vision},
  pages={19628--19637},
  year={2025}
}

@inproceedings{vincent2008extracting,
  title={Extracting and composing robust features with denoising autoencoders},
  author={Vincent, Pascal and Larochelle, Hugo and Bengio, Yoshua and Manzagol, Pierre-Antoine},
  booktitle={Proceedings of the 25th international conference on Machine learning},
  pages={1096--1103},
  year={2008}
}

@article{ho2020denoising,
  title={Denoising diffusion probabilistic models},
  author={Ho, Jonathan and Jain, Ajay and Abbeel, Pieter},
  journal={Advances in neural information processing systems},
  volume={33},
  pages={6840--6851},
  year={2020}
}

@article{heusel2017gans,
  title={Gans trained by a two time-scale update rule converge to a local nash equilibrium},
  author={Heusel, Martin and Ramsauer, Hubert and Unterthiner, Thomas and Nessler, Bernhard and Hochreiter, Sepp},
  journal={Advances in neural information processing systems},
  volume={30},
  year={2017}
}

@article{russakovsky2015imagenet,
  title={Imagenet large scale visual recognition challenge},
  author={Russakovsky, Olga and Deng, Jia and Su, Hao and Krause, Jonathan and Satheesh, Sanjeev and Ma, Sean and Huang, Zhiheng and Karpathy, Andrej and Khosla, Aditya and Bernstein, Michael and others},
  journal={International journal of computer vision},
  volume={115},
  number={3},
  pages={211--252},
  year={2015},
  publisher={Springer}
}

@inproceedings{zhang2018unreasonable,
  title={The unreasonable effectiveness of deep features as a perceptual metric},
  author={Zhang, Richard and Isola, Phillip and Efros, Alexei A and Shechtman, Eli and Wang, Oliver},
  booktitle={Proceedings of the IEEE conference on computer vision and pattern recognition},
  pages={586--595},
  year={2018}
}

@article{goodfellow2014generative,
  title={Generative adversarial nets},
  author={Goodfellow, Ian J and Pouget-Abadie, Jean and Mirza, Mehdi and Xu, Bing and Warde-Farley, David and Ozair, Sherjil and Courville, Aaron and Bengio, Yoshua},
  journal={Advances in neural information processing systems},
  volume={27},
  year={2014}
}

@article{wang2004image,
  title={Image quality assessment: from error visibility to structural similarity},
  author={Wang, Zhou and Bovik, Alan C and Sheikh, Hamid R and Simoncelli, Eero P},
  journal={IEEE transactions on image processing},
  volume={13},
  number={4},
  pages={600--612},
  year={2004},
  publisher={IEEE}
}

@article{salimans2016improved,
  title={Improved techniques for training gans},
  author={Salimans, Tim and Goodfellow, Ian and Zaremba, Wojciech and Cheung, Vicki and Radford, Alec and Chen, Xi},
  journal={Advances in neural information processing systems},
  volume={29},
  year={2016}
}

@article{tong2026beyond,
  title={Beyond Language Modeling: An Exploration of Multimodal Pretraining},
  author={Tong, Shengbang and Fan, David and Nguyen, John and Brown, Ellis and Zhou, Gaoyue and Qian, Shengyi and Zheng, Boyang and Vallaeys, Th{\'e}ophane and Han, Junlin and Fergus, Rob and others},
  journal={arXiv preprint arXiv:2603.03276},
  year={2026}
}

\end{document}